\documentclass[a4paper, 10 pt, conference]{ieeeconf}
\IEEEoverridecommandlockouts
\IEEEoverridecommandlockouts                              
                                                          
\usepackage{cite}
\usepackage{balance}
\usepackage{hyperref}
\usepackage{balance}
\usepackage{tabularx}
\usepackage{makecell}

\usepackage{amsmath,amssymb,amsfonts}

\usepackage[vlined,algoruled,commentsnumbered]{algorithm2e}
\usepackage{graphicx}
\usepackage{textcomp}
\usepackage{footmisc}
\usepackage{subcaption}

\usepackage[table,xcdraw]{xcolor}
\usepackage[utf8]{inputenc} 
\usepackage[T1]{fontenc} 
\def\BibTeX{{\rm B\kern-.05em{\sc i\kern-.025em b}\kern-.08em
    T\kern-.1667em\lower.7ex\hbox{E}\kern-.125emX}}

\SetKwFor{Times}{}{times do}{end}

\let\oldnl\nl
\newcommand{\nonl}{\renewcommand{\nl}{\let\nl\oldnl}}

\newlength\lenKwIn

\newlength\lenKwOut

\begin{document}

\title{\LARGE \bf
Human Intention Recognition for Human Aware Planning in Integrated Warehouse Systems}

\author{Tomislav Petkovi\'c$^{1}$, Jakub Hvězda$^{2}$, Tomáš Rybecký$^{2}$, Ivan Markovi\'c$^{1}$, \\ Miroslav Kulich$^{2}$, Libor Přeučil$^{2}$,  Ivan Petrovi\'c$^{1}$
\thanks{$^{1}$ Tomislav Petkovi\'c, Ivan Markovi\'c and Ivan Petrovi\'c are with the University of Zagreb Faculty of Electrical Engineering and Computing,
 Department of Control and Computer Engineering,
 Laboratory for Autonomous Systems and Mobile Robotics,
 Unska 3, HR-10000, Zagreb, Croatia
        {\tt\small (email: petkovic@fer.hr, ivan.markovic@fer.hr, ivan.petrovic@fer.hr)}.}%
\thanks{$^{2}$ Jakub Hvězda, Tomáš Rybecký, Miroslav Kulich and Libor Přeučil are with the Czech Institute of Informatics, Robotics and Cybernetics, Czech Technical University in Prague, 160 00 Prague, Czech Republic
       {\tt\small (email: hvezdjak@fel.cvut.cz, rybectom@fel.cvut.cz miroslav.kulich@cvut.cz, preucil@ciirc.cvut.cz)}.}%
}

\maketitle

\begin{abstract}

With the substantial growth of logistics businesses the need for larger and more automated warehouses increases, thus  giving rise to fully robotized shop-floors with mobile robots in charge of transporting and distributing goods.
However, even in fully automatized warehouse systems the need for human intervention frequently arises, whether because of maintenance or because of fulfilling specific orders, thus bringing mobile robots and humans ever closer in an  integrated warehouse environment.
In order to ensure smooth and efficient operation of such a warehouse, paths of both robots and humans need to be carefully planned; however, due to the possibility of humans deviating from the assigned path, this becomes an even more challenging task.
Given that, the supervising system should be able to recognize human intentions and its alternative paths in real-time.
In this paper, we propose a framework for human deviation detection and intention recognition which outputs the most probable paths of the humans workers and the planner that acts accordingly by replanning for robots to move out of the human's path.
Experimental results demonstrate that the proposed framework increases total number of deliveries, especially human deliveries, and reduces human-robot encounters.
\end{abstract}


\section{Introduction} \label{sec:intro}

\begin{figure}[!t]
  \centering
  \includegraphics[width=\linewidth]{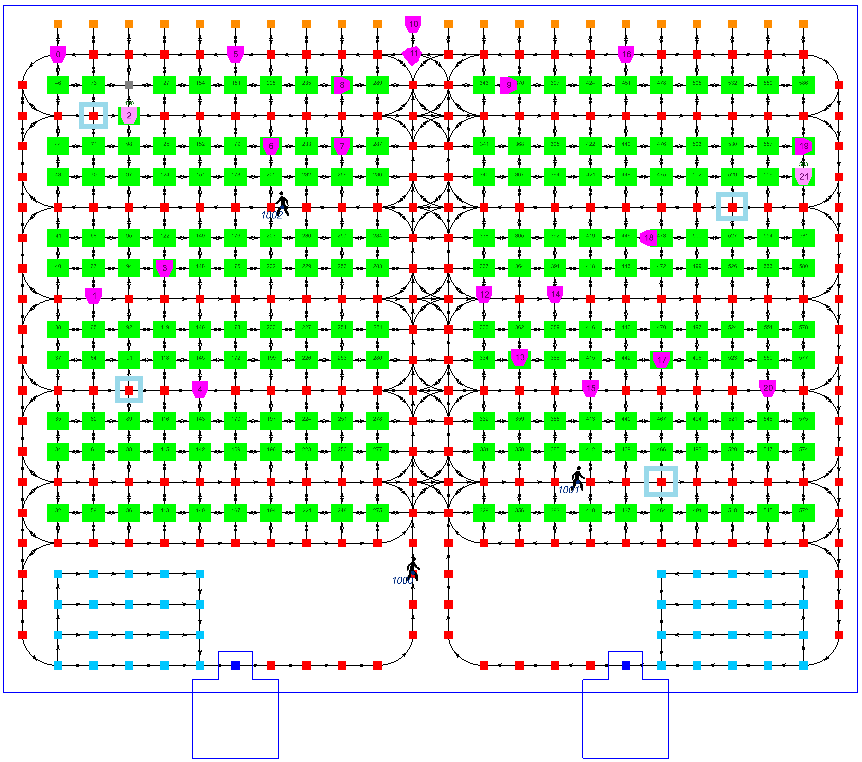}
  \caption{The developed flexible warehouse simulator. Mobile robots (pink) do a predefined set of tasks such as carrying storage racks (green) while moving on the ground nodes (red). Three human workers move freely between the storage racks picking goods at specified warehouse locations or doing maintenance work.}
  \label{fig:warehouse}
  \vspace{-0.6cm}
  \end{figure}

With the considerable global expansion of e-commerce in the past years a strong demand for efficient automation in the warehouse industry is following suit.
Given that, rapid processing of incoming orders and nonstop warehouse operation have become a paramount topic for companies having automated warehouses at the core of their businesses, e.g. Amazon and Swisslog, as well as for third-party logistics companies.
Even though complete warehouse automation with a considerable number of mobile robotic platforms is at the heart of such systems, the need for human intervention and collaboration with the robots still exists; indeed, it can even be beneficial for productivity and competitiveness \cite{IFR2017}.
A simulated example of a typical automated flexible warehouse can be seen in Fig.~\ref{fig:warehouse}.
It consists of large storage racks, packed with goods that are carried by a fleet of mobile robots to designated picking stations, where humans pick goods, pack them, and forward further for shipment.
Furthermore, humans can enter the warehouse shopfloor and move freely among the robots.
However, human workers do not always behave in a deterministic and prescribed fashion which can affect the carefully orchestrated coordination devised by the robot fleet management system (FMS); hence, to adapt to such perturbations, a human intention recognition (HIR) system is needed.

The task of the FMS, which carries out all the robot planning tasks, is to find trajectories between a pair of nodes in the warehouse while taking into account the plans of other robots.
The problem of trajectory planning and motion coordination is a well studied problem \cite{Parker2009}, and
since then many approaches have been introduced often based on the classical single-robot planning approaches \cite{Ryan2008,Latombe1991,Lavalle1998, phillips2011sipp} that provide completeness or even optimality; however, they are not practical for warehouse environments due to their large computational complexity.
To counter this issue, another category of sub-optimal planning algorithms has been introduced \cite{terMors2010Carp,drrt,peasgood08,Cap2015, hvvezda2019improved}, among which the context aware route planning (CARP) algorithm, in particular, provides good quality solutions in a warehouse environment with low path computation time and is easily extendable with warehouse specific constraints.
The main drawback of the original algorithm was its reliance on the ordering of the agents to be planned.
Several heuristic approaches have been introduced in \cite{terMorsHeuristics} to improve the properties of the algorithm.
However, in the warehouse environment the tasks are not known all at the same time but are given sequentially.

As stated earlier, to ensure warehouse operation efficiency, FMS needs to be complemented with HIR, thus effectively enabling a \emph{human aware planner} (HAP).
Given that, when a deviation has happened, HAP should be notified and assisted with the estimation of paths the human might follow -- this becomes possible if worker intentions are accurately recognized.
In the current paper, our approach to HIR relies on the \emph{Bayesian theory of mind} \cite{Baker2014a}, where authors introduced a model for estimating student desires to eat at a particular food-truck by observing their motion.
Examples of models using Markov decision processes can be found in \cite{Bandyopadhyay2013}, where authors proposed a framework for estimating pedestrian intention to cross the road, and in \cite{Lin2014}, where authors proposed a  gesture recognition framework for robot assisted coffee serving.
Learning based methods for trajectory estimation or HIR using motion pattern learning have also been studied in \cite{alahi2016social}; however, in \cite{Vasquez2009} this approach was criticized where authors emphasize that such techniques operate offline and imply that at least one example of every possible motion pattern is contained in the learning data set, which often does not hold in practice.
They propose using growing hidden Markov models for predicting human motion, a problem which we consider dual to the human intention estimation in the warehouse domain.
In \cite{Han2013} a thorough review is of human intention recognition is given emphasizing also its potential applications in decision making theory, and in \cite{rudenko2019human} authors give overview of human motion prediction methods.

In this paper we propose a robot fleet management system endowed with human worker intention recognition.
The HIR module is based on the Bayesian theory of mind giving a prediction of the human trajectory, should it deviate from the assigned path, and the proposed system is capable of taking into account the human deviations without the necessity to cease the operations of all the robots during human presence inside the warehouse.
The HAP reacts by moving the robots out of the human's way or simply stopping the robots.
Even though we assume that robots are equipped with a safety system, human aware planning can reduce the chance of robots driving close to humans thus lowering the number of times the safety system is triggered.
In conjunction, this leads to more efficient operation of the warehouse and hypothetically less stress on the human workers (note that load carrying robots can weigh up to 1000 kg).
We have benchmarked the proposed HIR framework with other methods and recorded increase in number of human deliveries by 207\%, increase in total deliveries by 28\% and reduction of human-robot encounters by 91\%.
\vspace{1.3cm}
\section{Proposed human aware robot fleet\\ management system} \label{sec:intention}

\begin{figure}[!t]
\centering
\includegraphics[clip, trim = 5.4cm 20cm 6cm 3.6cm, width=0.9\linewidth]{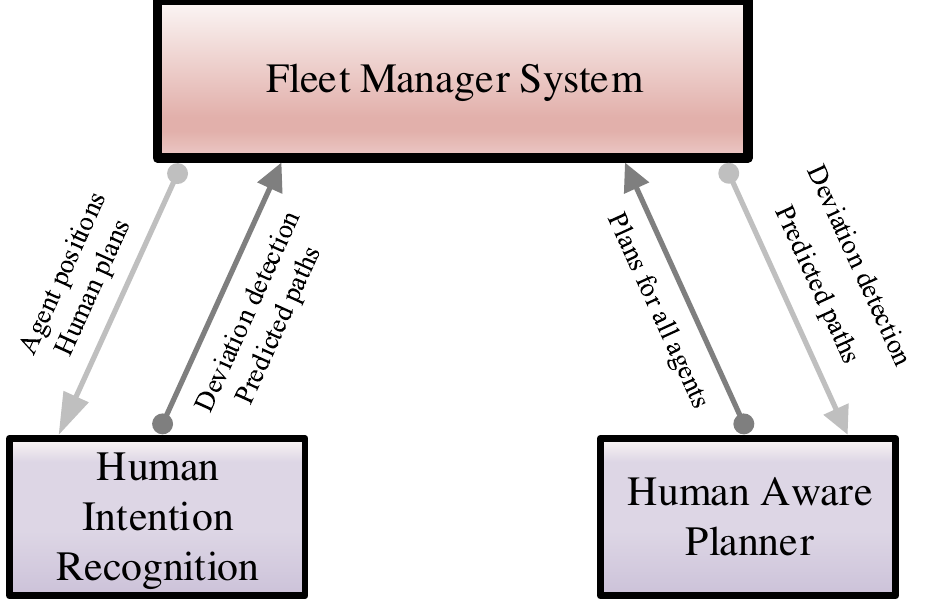}
\caption{Architecture of the proposed system. HIR and HAP are implemented as separate threads of the FMS. While the FMS is responsible for entire warehouse management, we have highlighted only the data flow which is described in the scope of this paper.}
\label{fig:system_architecture}
\vspace{-0.7cm}
\end{figure}

One of the main problems warehouse management currently faces is that most issues inside the warehouse shopfloor, e.g., a robot malfunction or goods falling on the floor storage racks, require human attention.
Furthermore, sometimes it is more efficient for a human worker to carry out the picking of goods if their distribution in the warehouse is too disparate.
Given that, there are many times a human intervention in the robotized warehouse is needed, and if this is not carried out in a planner manner it can lead to interruption of the whole warehouse operation, ultimately exacerbating the issue we were trying to solve in the first place.

The proposed system architecture is shown in Fig.~\ref{fig:system_architecture}.
FMS is in charge of planning paths for all the robots and humans as well (agents in short).
The HIR module serves as part of the FMS with the task of assisting the HAP.
HIR has at its disposal information about the position of all the agents in the warehouse, as well as planned paths of human workers.
While the robots are controlled by FMS and always follow the paths given to them, human workers can deviate from their paths for a number of reasons.
Because of that, supervising system can not assume that worker will always follow path that the FMS gives to them.
The job of the HIR module is to detect human deviations from the planned path and estimated the predicted trajectories, based on which the HAP will produce an updated plan for all the agents.
To efficiently assist the HAP, HIR should ascertain (i) if a worker deviated from the original path and (ii) where that particular worker is going to.
Given that, output of the HIR module is a logical flag indicating if there is a worker deviation, followed by a set of probable paths.
The HAP then reacts with the method described in Section~\ref{planning}.
In the sequel, we describe each of the proposed system's components in details.

\subsection{The warehouse simulator}

The simulated warehouse shown in Fig.~\ref{fig:warehouse} is organized into connected nodes and resembles faithfully the software architecture used by an FMS of a true robotized warehouse systems.
Each node can either be occupied by a robot, storage rack or a human.
Unloaded robots can move across all free nodes, while loaded robots cannot enter a node already containing a storage rack.
As incoming orders arrive, the robot fleet management system coordinates all the robots so that storage racks containing ordered goods are delivered to picking stations, then returned to a free storage node (not necessarily the same as the starting one), and the robots that are idle are sent to the charging nodes.
All this needs to be carried out in an efficient manner ensuring continuous operation of the whole warehouse.
This planning task is quite challenging since the number of robots can range from 50 to 800, and it gets an additionally layer of complexity by having to account for human workers in the area.
Note that we assume that robots are equipped with a safety system ensuring that the robots will stop if they come to a range that is too close to the human worker.
The multi-robot warehouse simulator was originally presented in \cite{hvvezda2018context}, while for the current paper we have extended the simulator with the ability to include human worker plan deviation.

\subsection{Human intention recognition}

\subsubsection{Human Deviation Detection} \label{deviation}

\begin{figure}[!t]
\centering
\includegraphics[clip, trim = 4cm 19.75cm 2cm 2.7cm, width=0.9\linewidth]{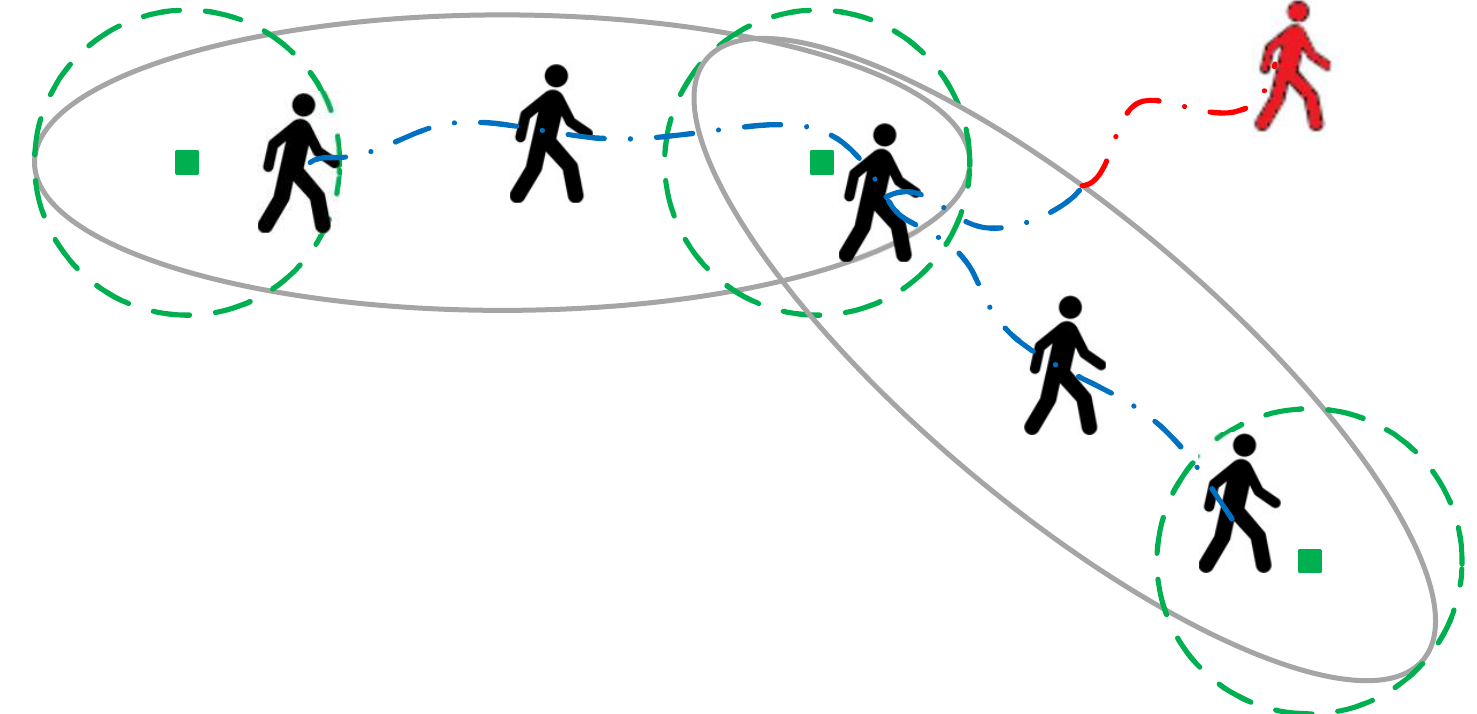}
\caption{The {allowed deviation area} is depicted by a grey ellipse defined by the {current node} and {next node} as focal point (green). Once the worker enters the green circle surrounding the {next node}, it becomes worker's {current node} and its successor in the path sequence becomes worker's {next node}. The example of human path which follows the plan is given with blue color while the path deviating from the plan is red.}
\label{fig:deviation_detection}
\vspace{-0.6cm}
\end{figure}

Each worker that enters the warehouse has a predefined path that consists of a sequence of ground nodes shown in Fig.~\ref{fig:warehouse}.
We designate the first node of the worker path as the \textit{current node}, and the second node of the path as the \textit{next node}.
Every time the distance between the worker and the next node in the path is less than $r = 0.25$\,m, that node becomes worker's \textit{current node} and its successor in the path sequence becomes worker's \textit{next node}, until the end of the path is reached.
Human beings usually do not walk in a perfect straight line \cite{bauby2000active}, but swing laterally while moving forward.
Given that, we allow deviation from the path defined by the \textit{allowed deviation area} that is described by an ellipse having focal points in the {current node} and {next node}, while the major axis is defined as the Euclidean norm between the focal points increased by $2r$.
The {allowed deviation area} is shown in Fig.~\ref{fig:deviation_detection} and the worker is considered to be deviating from the path if it has been outside of the area for at least 4 consecutive cycles.

\subsubsection{Human Path Prediction} \label{subsec:human_path_prediction}

If a worker is detected to be deviating from its predefined path, it is necessary to estimate its future path.
In this section we propose a human path prediction method relying on our recent result \cite{petkovic2019human}, where we proposed a human intention estimation method based on hidden Markov model (HMM) motion validation.
We describe it here briefly and direct the reader to the original paper for details.

We assume that there is a finite number of possible goal locations, which are usually in front of the storage racks of interest.
For current experiments, we selected four auxiliary goals, each one in a different corner of the warehouse and their locations are labeled with turquoise rectangles shown in Fig.~\ref{fig:warehouse}.
It is important to emphasize that the worker is not required to go to the predefined goals, but they do serve as starting points for the proposed algorithm.
Also, the last node of the human's path provided by the planner is also considered a goal location.

Before the simulation starts, we calculate the distance between all the ground nodes using the D$^*$ algorithm \cite{Stentz1997} and save it in a distance matrix $\textbf{F}$.
In case of a robot blocking the edge between two nodes during the experiment, we discard that edge from the graph and recalculate the distance matrix $\textbf{F}$.
Because we use road nodes, the search space is reduced significantly and the recalculation can be made in \textit{3.456}\, ms\footnote[1]{Configuration used for testing: Intel\textregistered Core\texttrademark  i7-7700HQ CPU @ 2.80GHz×8 with 15,5 GiB memory} for 228 nodes and 348 edges of the warehouse road graph.
For comparison, if we used a grid map representation with the precision of 10\,cm, the recalculation would be done on approximately $2\times10^{7}$ nodes and $1.5\times10^{8}$ edges which would make the recalculation time larger than one minute, thus rendering it too long for real-time application.

Each time a worker makes a significant displacement, we update its predefined goals intention estimate using a scaled down version of the algorithm proposed in \cite{petkovic2019human}.
We associate the position of the worker with the observable nodes by forming a so-called \emph{association vector} $\textbf{c}$.
The closer the human is to the node, the larger the value of the vector $\textbf{c}$.
By multiplying $\textbf{c}$ and $\textbf{F}$, and by isolating the goal nodes, we obtain a modulated distance vector $\textbf{d}$ of dimension $g$, where $g$ is the number of goals.
We also calculate the alternative association vector $\textbf{c'}$ of the positions the worker might have gone to, if it moved the same distance from the last observation; we also calculate the corresponding modulated distance vector $\textbf{d'}$.
By comparing values of the vector $\textbf{d}$ with values of each $\textbf{d'}$ that we collect in matrix $\textbf{D}$, we calculate the observation vector $\textbf{o}$ via element-wise division:
\vspace{-0mm}
\begin{equation}
\textbf{o} = \frac{\underset{1\leq i \leq n}\max{\textbf{D}_{ij}}-\textbf{d}}{\underset{1\leq i \leq n}\max{\textbf{D}_{ij}}-\underset{1\leq i \leq n}\min{\textbf{D}_{ij}}}.
\label{eq:validation}
\end{equation}
If the worker is moving towards a goal, the corresponding value of \textbf{o} will be close to unity, and if it is moving away from that goal, the corresponding value will gravitate to zero.
We record the observation history and process it with an HMM with $g+1$ states, one for each goal and one for the last node of human's predefined path.
We define the HMM's transition matrix \textbf{T} with  $\alpha=0.823$ on the diagonal and $\frac{1-\alpha}{g}$ otherwise.
We have obtained this parameter by learning on the recorded data with worker moving in the simulated warehouse without robots and minimizing the average displacement error \cite{pellegrini2009you}.
Using this formulation of \textbf{T} we allow the worker to change its mind of going towards any of the goals during the experiment.
Finally, we set initial probabilities of worker's intentions to $g^{-1}$ for each goal indicating that all of the goals are equally probable.
During the experiment we use the Viterbi algorithm \cite{Forney1973} to output probabilities of the worker going to each goal, which we consider as intention estimations.

After recording the probabilities of each goal we query if the probability of the worker going to the last node of the human's path is high enough by comparing it to the largest of the probabilities.
If their difference is less than the threshold of $0.25g$, we assume that the worker still might be going to the original goal and we report it to the HAP.
Otherwise, we find all goals with the probability higher than the threshold of $0.8g^{-1}$ and using the D$^*$ algorithm on warehouse nodes shown in Fig.~\ref{fig:warehouse}, we find the shortest path towards these goals on the road nodes.
These paths are then reported to the HAP.
For more detailed results on the path prediction we direct the reader to previous work \cite{petkovic2019humanicra}.


\subsection{Fleet Management System} \label{planning}

\subsubsection{Multi-robot route planning}

In this section we leverage the planning method we proposed in \cite{hvvezda2018context} that is based on the CARP algorithm \cite{terMors2010Carp}.
The original algorithm structures its map as a resource graph, where each resource has a corresponding timeline.
This timeline consists of free and occupied time windows, that indicate whether the resource is available in a given time interval or already taken by a different agent.
For each agent that needs planning, the algorithm then finds free time window on the corresponding resource to the start node of the agent and uses a modified A* algorithm to find the shortest path to free time window on the corresponding goal resource through expansion to the neighboring overlapping time windows.
The advantage of the used approach is that it takes into account only handful of the most influencing agents.
The algorithm aims to generate a trajectory for an agent $a_k$ while assuming that trajectories for \textit{k-$1$} agents are already planned.
This can lead to modification of those planned trajectories to accommodate the new agent.
The main idea is that the algorithm iteratively builds a set of agents whose trajectories affect the optimal trajectory of agent $a_k$ the most and replan this set of agents as well as agent $a_k$ in ordering that yields solution with the best global cost.

\begin{figure}[]
	\centering
	\begin{subfigure}[b]{0.55\columnwidth}
		\centering
		\includegraphics[width=0.9\linewidth]{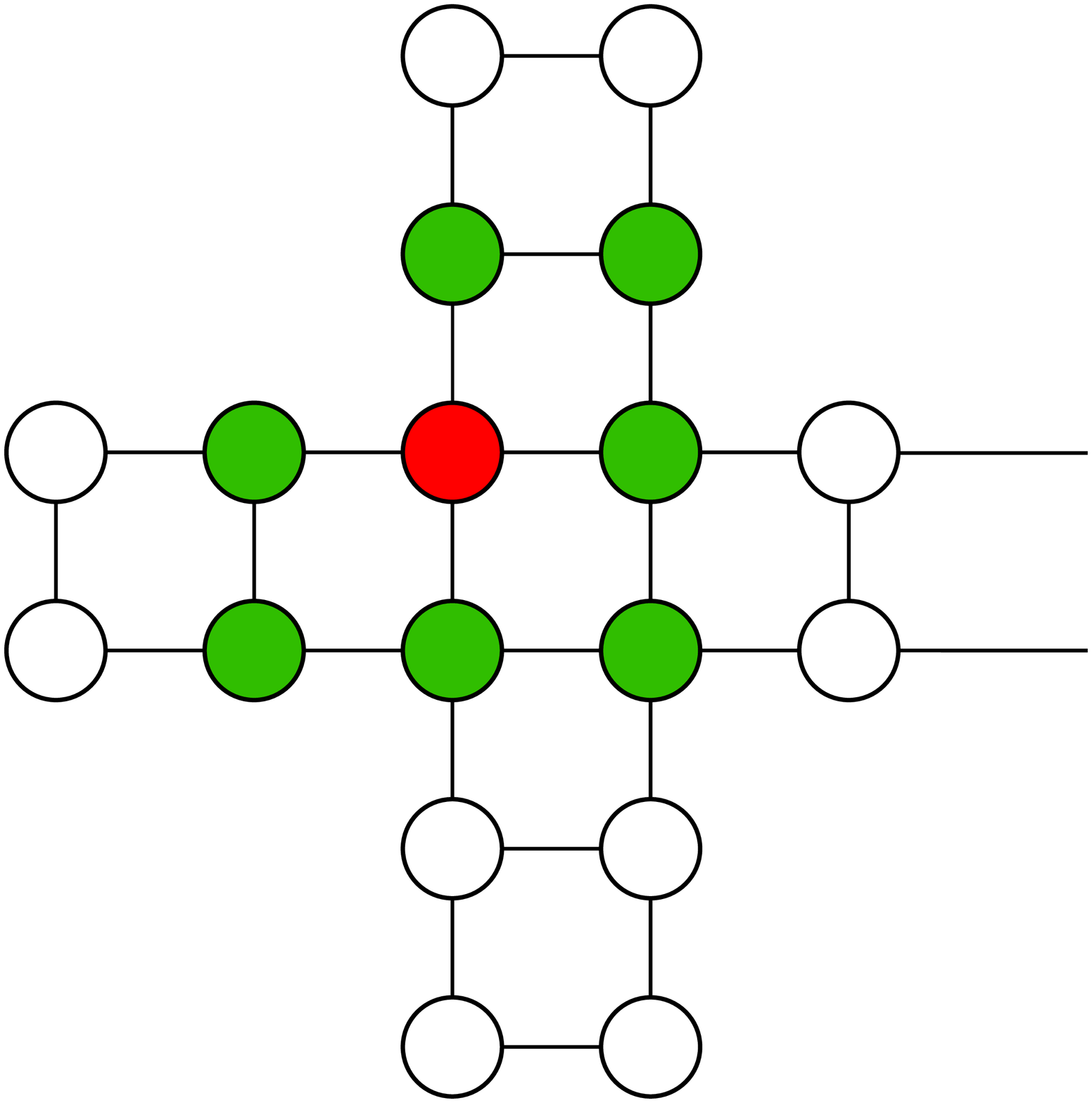}
		\caption{}
	
		\label{fig:Conflict2}
	\end{subfigure}%
	\hfill
	\begin{subfigure}[b]{0.4\columnwidth}
		\centering
		\includegraphics[width=0.6\linewidth]{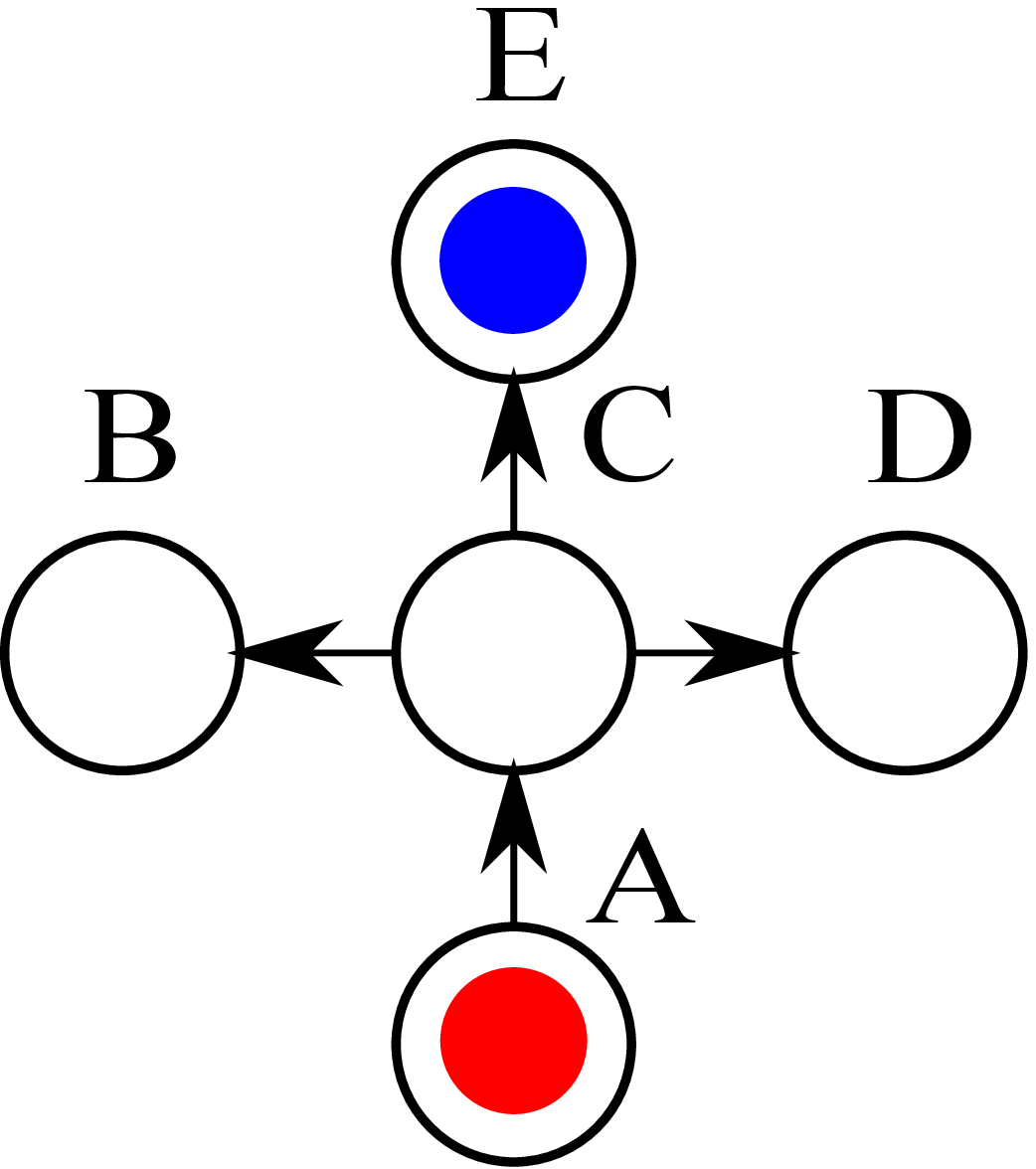}
		\vspace{3.5em}
		\caption{}
		
		\label{fig:Conflict}
	\end{subfigure}
	\caption{(a) An example of conflicting states. As long as the robot is standing on the \emph{Red} node, no other robot can stand on any of the \emph{Green} nodes. (b) An example of planning with conflicting states. \emph{Blue} robot wants to go to node \emph{B}, and \emph{Red} robot on node \emph{E}. The node \emph{C} is in conflict with nodes \emph{B}, \emph{D} and \emph{E}. This means that when  \emph{Blue} robot attempts to plan to node \emph{B}, the planning fails because even though the \emph{physical} timeline of node \emph{C} is free, the \emph{conflicting} timeline of node \emph{C} shows that it is occupied by the \emph{Red} robot.}
	\vspace{-0.5cm}
\end{figure}

Another type of constraint is the inclusion of the safety regions around the robot that differ in their radius and the interaction with robots.
There are three different safety regions defined, but only two of them are considered for planning:
\begin{itemize}
\item Safety region $1$: Robot stops its operation if it enters this radius
\item Safety region $2$: Planner avoids planning robot in this region
\item Safety region $3$: Robot must decrease its speed.
\end{itemize}
As the reader can notice, the safety regions $1$ and $2$ are identical for the planning algorithm, because the planner needs to avoid planning robots into them, and therefore the algorithm considers only regions $2$ and $3$.

To accommodate the constraints, the planning algorithm adds an additional timeline to each resource on the resource graph.
This timeline is referred to as \emph{conflicting}, with the main timeline is referred to as \emph{physical}.
The \emph{physical} timeline indicates when the resource is being physically occupied by an agent, while the \emph{conflicting} timeline indicates whether any resource that the current resource conflicts with is being physically occupied.
An example of conflicting states can be found in Fig.~\ref{fig:Conflict2}, while an example of planning with conflicting states is shown in Fig.~\ref{fig:Conflict}.
If these timelines were merged into one, the result would be the timeline where the free time windows are windows that the agent can move into without violating any constraints.

The safety regions for humans are handled similarly, i.e., by keeping timelines \emph{Safety region 2} and \emph{Safety region 3} for each resource and human present in the warehouse, corresponding to time windows that the human plan occupies.
Furthermore, the \emph{Safety region 2} timeline is added for each robot as well.
The \emph{Safety region 2} is used by merging the \emph{physical} and \emph{conflicting} timelines to obtain the final occupancy timeline of a resource during robot planning.
\emph{Safety region 3} occupancy timeline is used to check if the resource intersects the \textit{Safety region 3} during the computation of the time it takes the robot to cross a given resource.

\subsubsection{Robot path planning}

The proposed planner differentiates  five different states where a robot can be \textit{i)} going to pick up its rack assigned by the job, \textit{ii)} taking the rack to the start of the queue before the goal picking station, \textit{iii)} in the picking station queue, \textit{iv)} heading back to return it to its position or \textit{v)} heading back to its charging station. In addition to these planning states, the robot also has five internal states: \emph{Idle:} The robot is at the charging station, \emph{Busy:} The robot has a job assigned, currently working on its completion, \emph{Free:} No job assigned, returning to the charging station, \emph{Interrupted:} The state invoked by human action; the robot has been interrupted and needs to be replanned and \emph{Failed:} The robot failed to find a plan; switched to \emph{Interrupted} every few seconds.

\subsection{Human Aware Planner}
\subsubsection{Planning for humans}
We assume that the worker is equipped with a system, such as a hand-held screen or augmented reality glasses, that can navigate the human through the warehouse.
The planning is done in a similar manner as the robot planning; however, the human always takes precedence to robots in the planning process.
When the human planning starts, all the robots are interrupted.
Once all the robots have stopped, the planner attempts to find a path to the human goal destination, while considering stopped robots as obstacles and taking into consideration the \emph{Safety region 2} region where robots should not enter.
If such a path is found, it is returned by the system for the human to follow, and the system automatically replans the interrupted robots, while taking the human plan into account.
However, if such a path does not exist the system attempts to move the robots out of the way by first planning the human to the goal node, while not taking any of the robots into consideration.
The robots then attempt to plan their paths to the closest possible node to their current goal, that is not in conflict with the human path.
If the paths for all robots are found, it means that the evasive maneuver is possible and all the paths for human and robots are returned.
If none of these approaches succeed, then the system indicates that the planning was unsuccessful.
Moreover, to take into account the variance of human velocity, the system also plans the path for the human while taking into account the minimum and maximum velocity.
Each resource along the human path is taken for a time interval $w_i^r$ that starts at an entry time $t_{entry}^{fast}$, the time that would take the human to get to the goal if walking at maximum speed, and ends at the time $t_{exit}^{slow}$,  the time it would take the human to leave if walking at minimum speed.
Each time window $i$ for all resources $r$ in the path sequence is then $w_{i}^{r} = \langle t_{entry}^{fast}, t_{exit}^{slow} \rangle$.

\subsubsection{Planner reaction to the HIR input}
As described in Section~\ref{subsec:human_path_prediction}, once the human deviates from the planned path, the HIR determines paths to all the predefined goals whose probabilities exceed a given threshold.
Once the planner registers paths from the HIR module, it interrupts all the robots and finds the longest common path segment of the obtained paths.
This longest common path segment is then processed by the planner.
Notice that if the planner took into account all the paths, it would possibly block a large portion of the warehouse.
Once the segment that the planner will use for the human replanning is known, the system takes the first and last node as start $s$ and goal $t$ locations respectively .
The planner then attempts to find a plan from the start location $s$ to the goal location $t$ using the nodes that were present in the planning segment $\cal T$.
If this planning succeeds, the system tells the human that he deviated from his original path and informs him of the new plan.
However, if the planning is not successful, the system tells the human to stop immediately.

\section{Experimental Results}  \label{sec:results}

To demonstrate the system functionality, we have designed an experimental setup with several delivering scenarios.
All workers had a set of assignments, \textit{e.g.} picking or maintenance, that needed to be completed during the experiment.
We measured the average number of deliveries and human-robot encounters for cases when \textit{i)} humans deviate and planner reacts without HIR module (NHIR) and replanning is done only when human enters \emph{Safety region 1}, \textit{ii)} humans deviate and planner reacts using simple HIR module (SHIR) and \textit{iii)} humans deviate and planner reacts using proposed HIR module (PHIR).
The SHIR module outputs human path prediction on warehouse nodes with minimal change in heading assuming constant velocity.
\begin{figure}[t]
\centering
\includegraphics[clip, trim = 0.2cm 7.65cm 0cm 1cm, width=\linewidth]{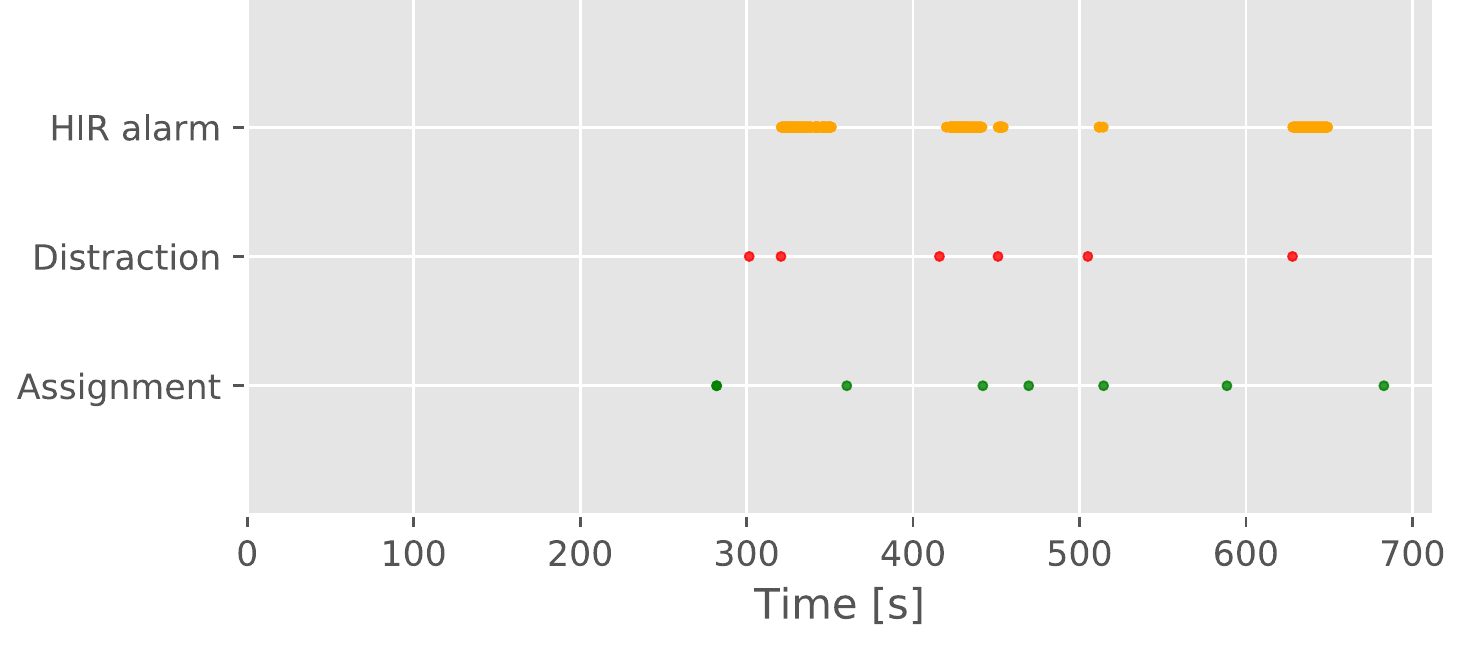}
\caption{Reaction of the HIR module to distractions.}
\label{fig:onlyHIR}
\vspace{-0.6cm}
\end{figure}

We have given humans assignments and simulated deviation multiple times during each experiment with random locations which were not known to the HIR module.
The HIR module reacts promptly with alarm and path prediction which is then handled by the HAP, which can be seen in Fig.~\ref{fig:onlyHIR}.
Once the path is accepted by the HAP and human starts following it, the HIR alarm is turned off.
\begin{table}[b]
\vspace{-0.6cm}
\centering
\begin{tabular}{l l l l l l}
     \Xhline{4\arrayrulewidth}
   & NHIR & SHIR & PHIR \\
 \hline
  \multicolumn{4}{c}{One Human} \\
    \hline
 Robot Deliveries            & 57.9  & 53.2  & \textbf{58.1} \\
 Human Deliveries  & 9.4 & 8.8  & \textbf{12.0}  \\
 Total Deliveries & 67.3 & 62.0 &\textbf{70.1} \\
 Human-Robot Encounters / min& 5.40   & \textbf{0.39}  & 0.49\\
 \hline
   \multicolumn{4}{c}{Three Humans} \\
    \hline
   Robot Deliveries  & 48.5  & 31.7  & \textbf{57.4} \\
 Human Deliveries  & 3.0 & 4.4  & \textbf{9.2}   \\
 Total Deliveries & 51.5 & 36.1 & \textbf{66.6} \\
 Human-Robot Encounters / min & 0.13  & 0.13  & 0.13  \\
     \Xhline{4\arrayrulewidth}
\end{tabular}
\caption{Average experimental results for ten experiments lasting 750 seconds.}
 \label{tbl:results}
\end{table}
We have conducted two experimental scenarios, the first with a single human worker and the second with three human workers.
The results of ten experiments for each scenario lasting 750\,s with unique job sets for all agents can be seen in  Table~\ref{tbl:results}.

Specifically, for the single human scenario we have achieved an increase of $28\%$ in human deliveries and $4\%$ in total deliveries, while for the three humans scenario, we have achieved increase of $18\%$ in robot deliveries, $207\%$ in human deliveries and $29\%$ in total deliveries.
Results suggest that correct prediction of human intention can improve warehouse throughput when integrated with a HAP, especially in cases when there are multiple humans operating in the warehouse at the same time.
Furthermore, an interesting side-effect of the proposed method is reduction of the number of human-robot encounters.
By reducing human-robot encounters, the system can hypothetically reduce discomfort and stress of human workers, since each close encounter with the robot triggers a robot safety stop (loaded warehouse robots can weigh close to 1000\,kg).
For the single human scenario the number of encounters was reduced by 91\%, while for the three humans scenario this number remained unchanged.
It would be interesting for future work to investigate the behavior of the HIR enhanced HAP with respect to the number of human-robot encounters and the increasing number of workers in the warehouse.

\section{Conclusion}  \label{sec:conclusion}

In this paper we have proposed a human intention recognition framework for human aware planning in integrated warehouse systems.
The framework detects humans deviating from assigned paths and based on an HMM approach identifies and outputs the most probable paths that human worker is about to take.
This information is fed to the human aware planner that is able to account for such deviations and replan the paths of warehouse robots so that warehouse efficiency is kept.
We have conducted experimental runs on an in-house developed simulator and demonstrated that with human intention recognition we increase the number of total deliveries, especially human deliveries, which for the scenario with three humans increased by 207\%.
Furthermore, the proposed method has the potential to also reduce the number of human-robot encounters, which decreased by 91\% in the single human scenario.

\section*{ACKNOWLEDGMENT}

This work has been supported from the European Union's Horizon 2020 research and innovation programme under grant
agreement No 688117 ``Safe human-robot interaction in logistic applications for highly flexible warehouses (SafeLog)''. The work of Jakub Hvězda was also supported by the Grant agency of the Czech Technical University in Prague, grant No. SGS18/206/OHK3/3T/37.


\bibliographystyle{IEEEtran}
\balance
\bibliography{library}

\end{document}